# Learning Neural Activations

**Fayyaz ul Amir Afsar Minhas[1], [*] and Amina Asif [2, 3]**

[1] Department of Computer Science, University of Warwick, Coventry CV4 7AL, UK

[2] Department of Computer and Information Sciences, Pakistan Institute of Engineering and Applied Sciences (PIEAS), PO Nilore, Islamabad, Pakistan.

[3] Department of Computer Science, National University of Computer and Emerging Sciences FAST-NU, Islamabad, Pakistan

[*] *corresponding author email: fayyaz.minhas@warwick.ac.uk;*

**Abstract**

An artificial neuron is modelled as a weighted summation followed by an activation function which determines its output. A wide variety of activation functions such as rectified linear units (ReLU), leaky-ReLU, Swish, MISH, etc. have been explored in the literature. In this short paper, we explore what happens when the activation function of each neuron in an artificial neural network is learned natively from data alone. This is achieved by modelling the activation function of each neuron as a small neural network whose weights are shared by all neurons in the original network. We list our primary findings in the conclusions section. The code for our analysis is available at: https://github.com/amina01/Learning-Neural-Activations.

## 1. Introduction

*"What is the optimal activation function in a neural network?"* is a question that is still being researched after more than 70 years since the first reported use of a binary activation function in the McColluch-Pitts perceptron model [1]. A number of researchers have attempted to answer this question by proposing different types of activation functions whose design is mostly motivated by some desirable criteria such as monotonicity, differentiability, non-saturation, bounded-support, etc. [2]–[6]. The question we pose in this paper is *"Can we learn the activation function of a neural network while training the neural network*? and if yes, then *what are the characteristics of such learned activation function*?".* Early work on this question has been carried out by Fausett [7] who devised a method to learn the slope of sigmoidal activation functions. However, this analysis was limited to changes in the slope of the activation only. A more generic approach was recently proposed by Ramachandran et al. which used an automatic search technique to discover SWISH activations [4]. Other approaches that involve searching for an activation function through error minimization include [8], [9]. Similar to these methods, we have analyzed what type of activation functions can be learned using empirical risk minimization while learning the overall neural network for a given problem. In contrast to previous work, our proposed approach is not motivated by the design of better activation functions that may give improved accuracy or reduced computational complexity. Rather, we simply want to analyze what activation functions are learned naturally if the only objective is empirical error reduction. Our approach is based on updating the activation function every time the prediction neural network is updated. In conclusion, we also give pointers on how the problem of learning activation functions from data can be formulated as a meta-learning problem.

Table 1- Some commonly used activation functions.

| Name and Reference | Formula for $g(z)$ |
|---|---|
| **Linear** | $z$ |
| **Rectified Linear (ReLU)** | $\max(0, z)$ |
| **Leaky-ReLU** | $\begin{cases} 0.01z \ for \ z < 0 \\ z \quad \ for \ z \geq 0 \end{cases}$ |
| **Sigmoid (or logistic)** | $1/(1 + e^{-z})$ |
| **Hyperbolic Tangent (tanh)** | $(e^{2z} - 1)/(e^{2z} + 1)$ |
| **Swish [4]** | $z \cdot sigmoid(z)$ |
| **MISH [6]** | $z \, tanh(ln(1 + e^z))$ |

## 2. Methods

### 2.1 Structural and Mathematical Formulation

Mathematically, for a given *d*-dimensional input $x \in \Re^d$, a single neuron with weights $w$ can be modelled as a weighted summation $z(x; w) = w^T x$ followed by an activation function $g(\cdot)$ as follows: $f(x) = g(w^T x)$. The activation function is typically set *a priori* by the designer of the neural network. Examples of some activation functions are given in Table 1. A predictive neural network model can be expressed as a mathematical function $F(x; \theta)$ with all learnable weight parameters of all its constituent neurons lumped into $\theta$. In case of supervised learning, the neuron network is learned using structural or empirical risk minimization which involve minimizing a loss function $l(F(x; \theta), y)$ that measures the extent of the error between the output of the neural network for a given input and its associated target.

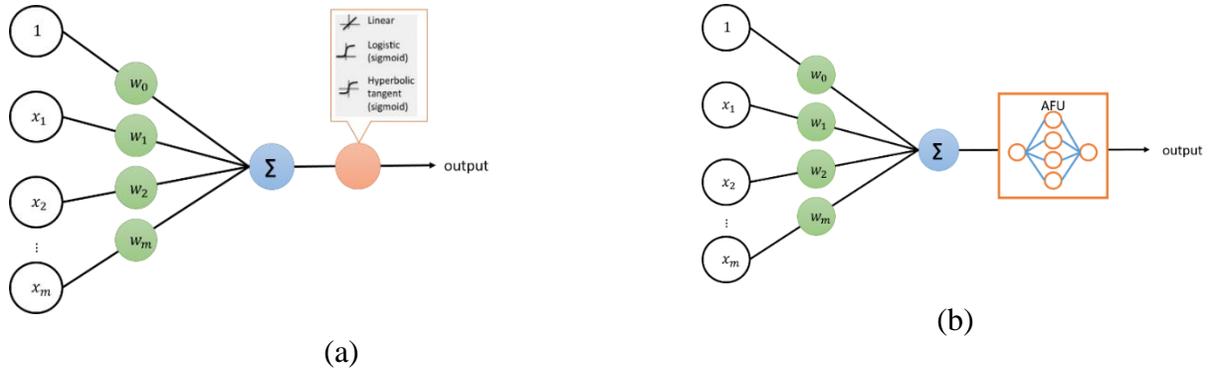

**Figure 1- (a) Illustration of a neuron with a pre-defined activation function. (b) Concept diagram of a neuron using learnable AFU for activation.**

In this work, we want to analyze what happens when the activation function of each neuron or a group of neurons in a neural network model is parametrized by a small neural network, henceforth referred to as an Activation Function Unit (AFU). Mathematically, the output of an AFU can be represented by $g(z) = G(z; \kappa)$. Here $G$ represents the neural architecture of the AFU and $\kappa$ are its learnable parameters. In this work, we have primarily explored single hidden layer neural networks as AFUs, i.e., functions of the parameterized form given by: $G(z; \kappa) = \sum_{i=1}^{N} \kappa_{1,i}^{w} \psi(\kappa_{0,i}^{w} z + \kappa_{0,i}^{b}) + \kappa_{1,1}^{b}$ (see Figure 1). Here $\psi(\cdot)$ is a base activation function and can be set to any of the canonical activation functions listed in Table 1 and $N$ is the number of hidden units in $G$. The weights and bias of the i$^{th}$ neuron in j$^{th}$ layer in the AFU are denoted by $\kappa_{j,i}^{w}$ and $\kappa_{j,i}^{b}$, respectively. The overall learning problem can now be framed as: $\min_{\theta; \kappa} l(F(x; \{\theta; \kappa\}), y)$. Note that AFU parameters can either be shared by all neurons in the neural network $F$, i.e., all neurons in the network apply the same AFU or each layer (and possibly each individual neuron) in $F$ can have their own AFU (see Figure 1). Since the AFUs are shared by multiple neurons in $F$, the increase in the overall number of learnable parameters is quite small: 3N+1 for the single hidden layer AFU with N neurons in the hidden layer given above.

## 2.2 Experimental Analysis

### 2.2.1 Toy Problem

We have first solved a simple two-dimensional two-class XOR-style classification problem with the proposed approach. The data comprises of a total of 2000 partially overlapping points drawn from four Gaussian distributions (500 points each, two per class) centered at (-1, -1), (+1, +1)

(positive class) and (-1, +1), (+1, -1) (negative class). The prediction neural network comprises four hidden layer neurons sharing an AFU with a single output layer neuron with linear activation. Hinge loss function $l(F(x;\boldsymbol{\theta}), y) = \max(0, 1 - yF(x;\boldsymbol{\theta}))$ is used together with adaptive moment estimation (Adam) [10] for gradient based optimization. We have experimented with different number of neurons (8,128) in the AFU hidden layer as well as different base activation functions (ReLU, Sigmoid). In terms of performance analysis, we show the classification boundary learned for the problem together with the activation function and activation outputs of individual neurons.

### 2.2.2 Smoothness Analysis

In order to analyze the smoothness characteristics of the AFU learned for the toy problem above, we have compared the prediction landscape for a randomly initialized neural network (5 layers with 10 neurons in each layer) using ReLU vs. AFU activations.

### 2.2.3 MNIST

For a more practical analysis, we have used the MNIST dataset [11] (60K training and 10K test examples). Here, the prediction neural network consists of two convolutional layers (32 3×3 filters with a 25% training drop-out in the first and 64 3×3 filters in the second) followed by two fully connected layers (128 neurons in the hidden layer with 50% training drop-out and 10 output layer neurons corresponding to the ten classes). Negative log likelihood is used as the loss function in this problem. An adaptive learning rate method (ADADelta) is used together with a step-wise learning rate scheduler which decreases the learning rate by 0.7 in each epoch starting from a base learning rate of 1.0. Here, we did two experiments: 1) same AFU across all neurons in all layers and 2) different AFU for different layers. The learned AFUs for these experiments together with the prediction accuracy using the widely used ReLU activation and AFU are compared for a fixed number of training epochs (10). We have also plotted the loss function across epochs for the proposed approach with ReLU activation.

### 2.2.4 CIFAR-10

We have also analyzed the proposed approach over the CIFAR-10 [12] benchmark dataset which comprises of 10 classes (airplane, dog, automobile, bird, cat, deer, frog, horse, ship, truck) each with 6000 32×32 images. We have used the MobileNetv2 [13] in this experiment. The neural network is trained over 50,000 training examples for 100 epochs using a batch size of 100 with Adam optimizer (learning rate of 0.001). Test performance is evaluated over 10,000 examples and the accuracy of the network with the same AFU across all layers is compared to using MISH [6] activation which gives ~86% accuracy.

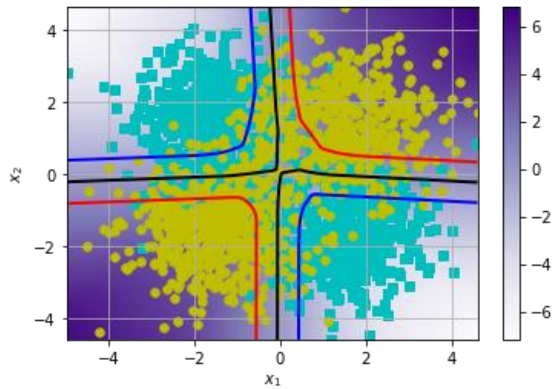

(a) Learned decision boundary (black solid line)

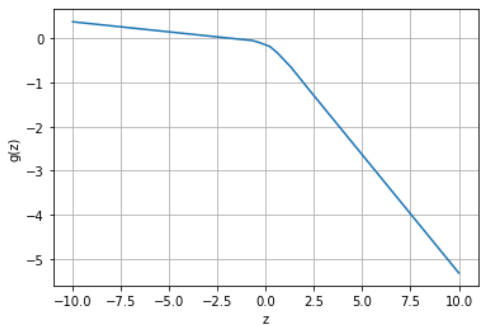 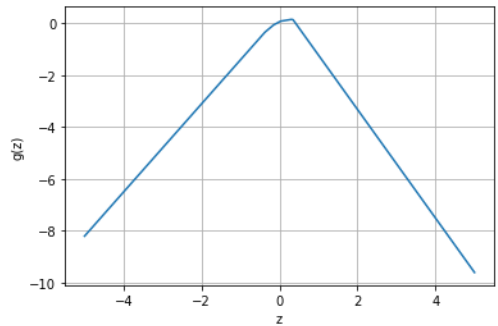

(b) AFU before training    (c) AFU after training

**Figure 2- Learnt decision boundary and AFUs for the toy problem.**

## 3. Results and Discussion

### 3.1 Toy Problem

For the XOR toy problem, the boundary produced by the prediction network is shown in Figure 2. The AFU for the neural network is plotted before and after training in the figure as well. It is interesting to see that the learned AFU is very different from the initial AFU which uses ReLU activations in its design. Furthermore, the AFU loosely resembles a radial basis function centered at zero. There was no improvement in accuracy using this AFU over and above native ReLU, sigmoid or hyperbolic tangent activation functions. The activations of individual neurons in the neural network are plotted in Figure 3. The learned AFU for this problem with using a sigmoid base activation function with 8 and 128 hidden layer neurons are shown in Figure 4 (a) and (b) respectively. It is interesting to observe the overall similarity of the learned AFU with sigmoidal base activations to the one obtained using ReLU activations. It is also interesting to note that the learned activation functions do not saturate and do not produce zeros which can result in a large number of "dead" neurons.

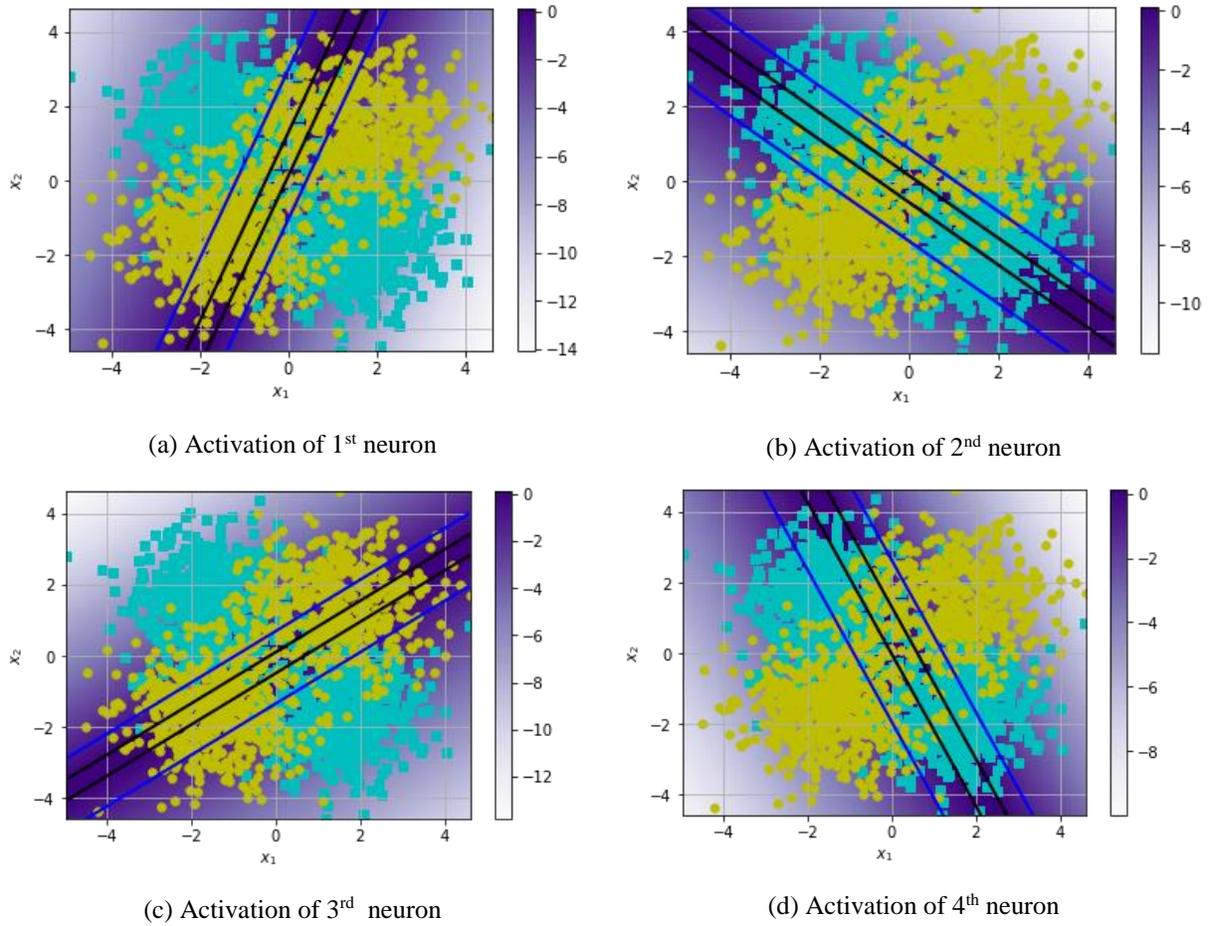

(a) Activation of 1st neuron

(b) Activation of 2nd neuron

(c) Activation of 3rd neuron

(d) Activation of 4th neuron

**Figure 3- Activation plots for 4 neurons in hidden layer of the neural network for toy problem.**

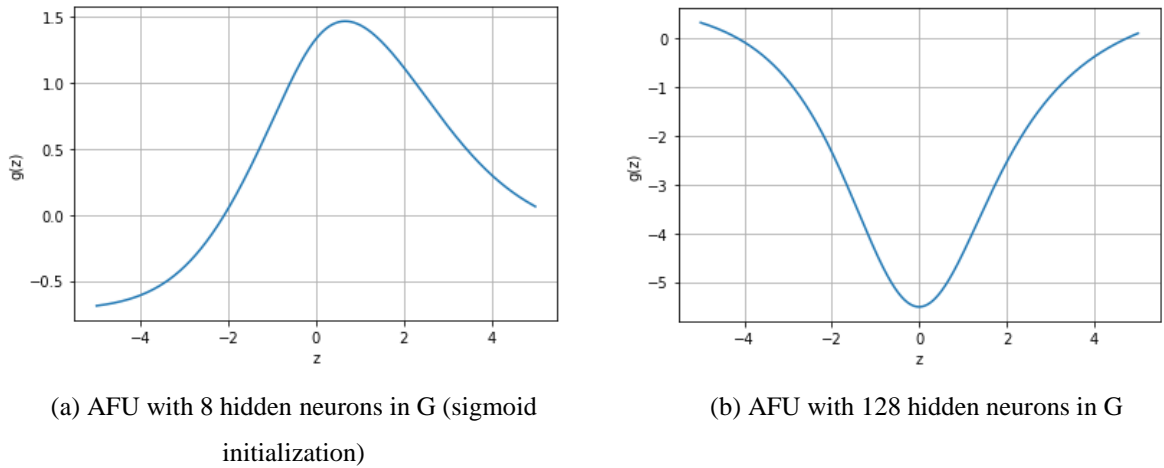

(a) AFU with 8 hidden neurons in G (sigmoid initialization)

(b) AFU with 128 hidden neurons in G

**Figure 4- AFUs learnt using 8 and 128 neurons in hidden layer with sigmoid as base activation.**

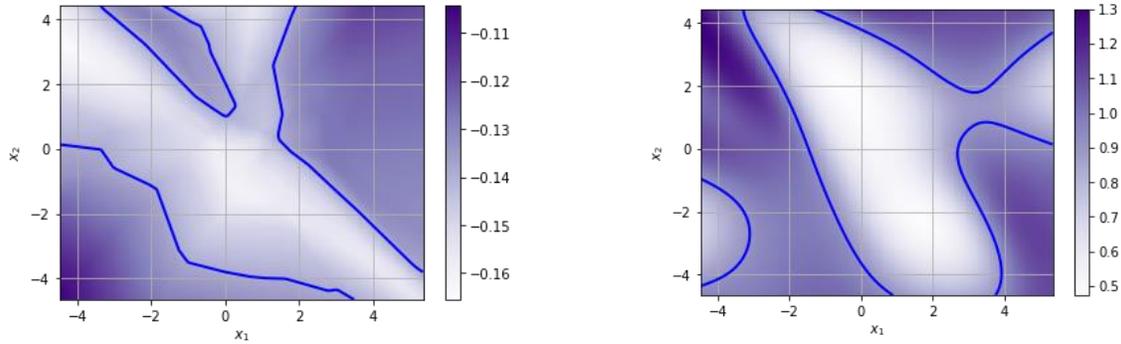

(a) Prediction scores for a randomly initialized 5-layer network with ReLU

(b) Prediction scores for a randomly initialized 5-layer network with the learned AFU over the toy problem

**Figure 5- Comparison of smoothness between ReLU and AFU learnt using the proposed scheme.**

## 3.2 Smoothness Analysis

We have analyzed the smoothness of the learned AFU relative to ReLU over a 5-hidden layer randomly initialized network with 10 neurons in each layer in Figure 5. The results show that the learned AFU results in smoother transitions in comparison to the canonical ReLU network which indicates a possible and unproven regularization impact of learning activation functions.

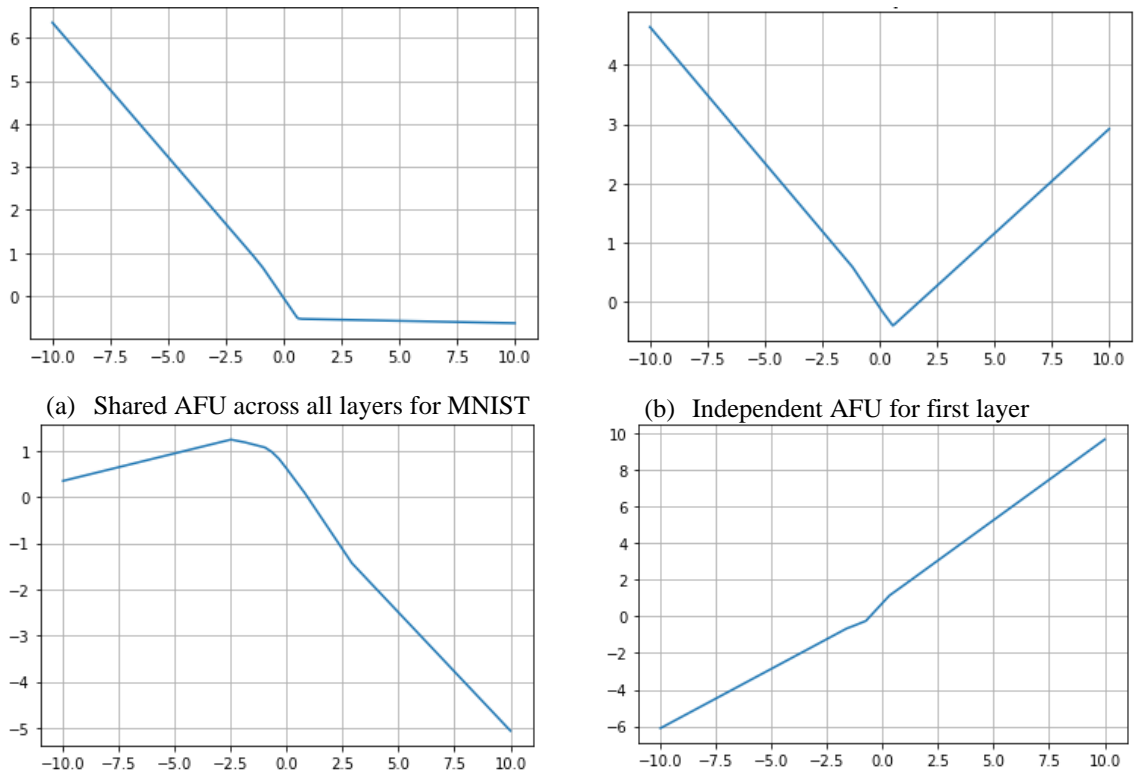

(a) Shared AFU across all layers for MNIST

(b) Independent AFU for first layer

(c) Independent AFU for 2$^{nd}$ layer

(d) Independent AFU for 3$^{rd}$ layer

**Figure 6- Learnt AFUs for MNIST experiment.**

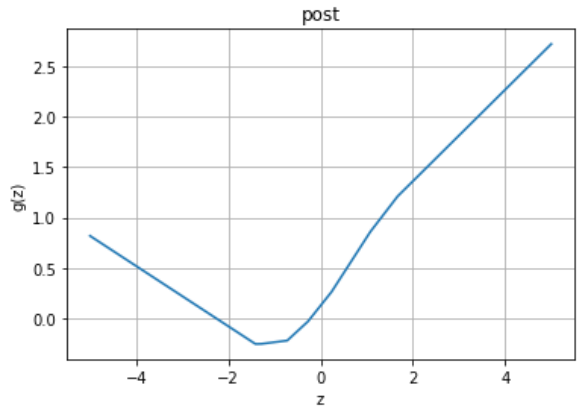

Figure 7- Learnt AFU for CIFAR-10 experiment.

### 3.3 MNIST

For the MNIST problem, we have obtained no significant improvement in prediction performance when using learned AFU over and above canonical activation functions such as ReLU and hyperbolic tangents are used (Accuracy of 99.2% was obtained for all). However, there is a significant difference between the learned AFU when the same AFU is shared across all layers and when different layers used different AFUs (see Figure 6). The same AFU learned for all layers resembles a leaky-ReLU. In contrast, when the AFUs are allowed to be different for different layers in the network, the resulting activation functions are indeed very different from each other. A V-shaped activation function similar to the one obtained for the toy problem is produced for the first layer whereas an inverted Leaky-ReLU like activation is produced for the second layer. A mostly linear activation is learned for the fully connected layer. This seems to indicate that learning activation functions can have a self-regularizing effect (unproven).

### 3.4 CIFAR-10

Similar to the MNIST analysis, no performance improvement is observed for the CIFAR10 dataset when using learned AFUs vs. canonical AFUs (e.g., MISH): Using MISH gives top-1 accuracy of 86.3% whereas AFU results in 85.1% (for the same mobile net v 2 with 100 epochs). However, it is interesting to see that the resulting learned AFU is similar in structure to MISH (Figure 7).

### 4. Conclusions and Future Work

In conclusion we have found that:

- Learned activation functions do not degrade performance and can possibly lead to minor performance gains. This has also been shown in the independent work by[9].
- Learning activation functions typically result in smooth non-saturating functions.

- The use of learned activation functions in deep neural network can produce smooth fitness landscapes possibly due to the over-parameterization of the optimization problem.
- Learned activation functions can have a possible regularization effect on learning. However, this effect needs to investigated further.
- Natively learned activation function can be very similar to well-known activation functions (such as Leaky-ReLU or MISH).
- Learned activation functions produce a V (or inverted V) type pattern which needs to be explored further.
- Different layers can learn vastly different activation functions.
- In our experience, more linear-like activations are learned towards the final layers of a deep neural network. As a consequence, it may be advisable to have fixed initial linear activation functions in the final layers when initializing the neural network and gradually making them more and more non-linear similar to ideas underlying curriculum learning [14].
- Based on analysis of the structure activation functions learned by our approach we conclude that MISH activation can possibly be further improved slightly by addition of a parameter that controls its slop for in the negative quadrants ($z<0$).
- Activation functions learned in our independent work are very similar in structure to the ones produced by [9] despite our proposed scheme being relatively naïve in comparison.
- It would be definitely interesting to explore what types of performance gains are obtained for a machine learning problem when using activation functions learned over different but related classification problem.
- Learning better activation functions can be modelled as meta-learning problem similar to the work [15], [16]. Here, the change in the loss function of a neural network in a fixed number of steps using a given AFU can be used as a meta-loss function to optimize the AFU resulting in activation functions that guarantee faster convergence.